\documentclass[letterpaper,journal]{IEEEtran}

\usepackage{amsmath,amsfonts,bm}

\def\eqref#1{equation~\ref{#1}}

\def\1{\bm{1}}

\DeclareMathAlphabet{\mathsfit}{\encodingdefault}{\sfdefault}{m}{sl}
\SetMathAlphabet{\mathsfit}{bold}{\encodingdefault}{\sfdefault}{bx}{n}

\usepackage{graphicx} 
\usepackage{cite}     
\usepackage{microtype}
\usepackage{fewerfloatpages}

\usepackage{hyperref}
\usepackage{url}
\usepackage{booktabs}
\usepackage{tabularx}
\usepackage[table]{xcolor}
\usepackage{adjustbox}
\usepackage{array}
\usepackage[T1]{fontenc}
\usepackage{xcolor}
\usepackage{amsmath}
\usepackage{multirow}
\usepackage[table]{xcolor}
\usepackage[most]{tcolorbox}
\renewcommand{\arraystretch}{1.25}

\newcommand{\drop}[1]{\,\textcolor{green!50!black}{\scriptsize$\downarrow$#1\%}}
\newcommand{\nodrop}{\,\textcolor{green!50!black}{\scriptsize 00.0\%}}

\newcommand{\kvmem}[2]{%
  \textcolor{blue!70!black}{#1}\,/\,%
  \textcolor{brown!80!black}{#2}%
}

\newcommand{\kvsaving}[2]{
  \textcolor{blue!70!black}{#1}\,/\,%
  \textcolor{brown!80!black}{#2}%
}

\begin{document}

\title{Rethinking Visual Token Retention in Multimodal Reasoning}

\author{Fawaz Sammani and Nikos Deligiannis\\Vrije Universiteit Brussel, ETRO Department}

\maketitle

\begin{abstract}
Autoregressive Large Vision-Language Models solve multimodal reasoning tasks by retaining hundreds or thousands of visual tokens, allowing every generated token to attend to the visual tokens at any time during the generation process. However, whether continuous access to these visual tokens is necessary remains poorly understood. In this work, we investigate how visual token influence evolves along two dimensions: generation time and transformer depth. Using long-form visual reasoning benchmarks and multiple model scales, we first analyze the attention dynamics across generation time and transformer depth. We identify a temporal and depth-wise structure in the use of visual information. Visual attention peaks during the initial generation steps and then rapidly declines as attention shifts toward previously generated reasoning tokens. Across depth, visual influence is concentrated in intermediate layers and becomes very weak in the final layers. We further verify these observations with causal intervention that removes all visual tokens across generation time and model depth. These interventions confirm that visual-token dependence in multimodal reasoning is neither temporally persistent nor uniformly distributed across transformer depth. Thus, retaining the complete visual context at every layer and decoding step may be unnecessary. Our experiments demonstrate that we can remove up to 59\% of tokens from the full input prompt across generation time and 11\% across all transformer layers, while incurring only a very small performance drop.  On video benchmarks, up to 65\% of the input prompt tokens can be removed during generation without affecting performance. 
\end{abstract}

\section{Introduction}
\IEEEPARstart{L}{arge} Vision-Language Models (LVLMs)\cite{Qwen2.5-VL,Qwen3-VL} have rapidly evolved into systems capable of solving multimodal mathematics, visual logic, diagram understanding, and other reasoning tasks. In these settings, a model must not only recognize visual content but also sustain a long chain of reasoning grounded in that content. Existing LVLMs typically support this process by inserting hundreds or thousands of visual tokens into the context of the Large Language Model (LLM) they contain, and retaining them throughout autoregressive generation. As a result, every generated token can, in principle, attend back to the original image tokens at every transformer layer. 

However, it remains unclear whether LVLMs actually use visual information in this way. Does a model repeatedly revisit the image throughout a long autoregressive reasoning trajectory? Are visual tokens equally important across the depth of the network during reasoning, or is their contribution concentrated in only a subset of layers? Mechanistically, these questions concern how perception and reasoning function inside LVLMs. Space-wise, retaining the full visual context increases the key–value cache memory at every decoding step, even when the model may no longer rely on the image. A fundamental question therefore arises: \textit{how does the influence of visual information evolve throughout autoregressive generation and model depth, and must visual tokens remain accessible during the entire reasoning process?} A follow-up question is then: \textit{what trade-off exists between removing all visual tokens and preserving model performance?}

Existing studies have shown that LVLM predictions can be strongly influenced by the LLM they contain, owing to the strong linguistic priors acquired during LLM pretraining \cite{pezeshki2021gradient,pmlr-v162-huang22e, parcalabescu2025do, Deng2025WordsOV,park2025generalizing,RevistingLin,Tong2024EyesWS}. Nevertheless, these works do not explain how visual dependancy and influence changes as a long autoregressive reasoning response unfolds, how the visual influence on reasoning is distributed across transformer layers, or whether visual information is repeatedly retrieved from the original image tokens during generation.  

\begin{figure}
    \centering
    \includegraphics[width=0.97\linewidth]{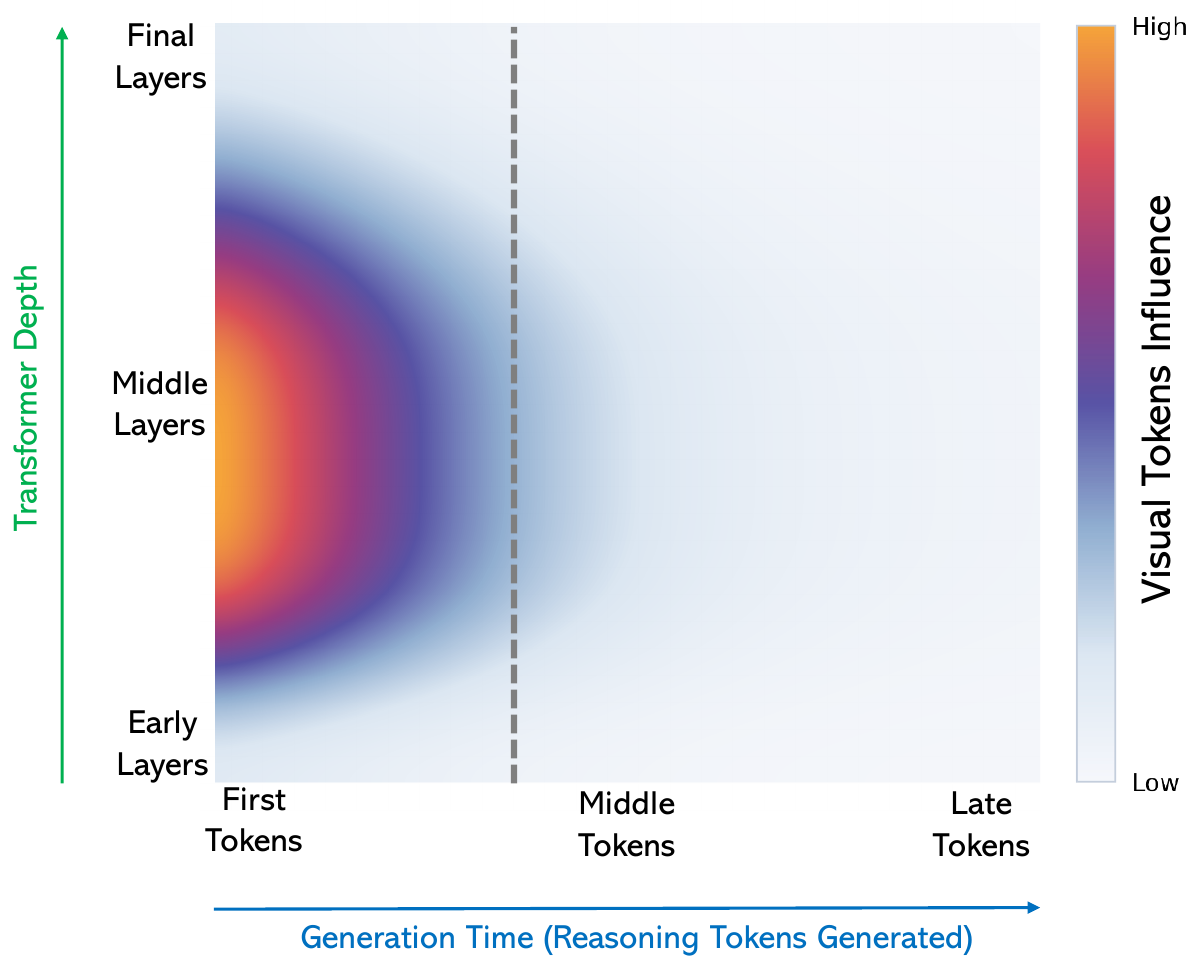}
    \caption{\textbf{Temporal (x-axis) and layer-wise (y-axis) dynamics of visual tokens influence in LVLM reasoning}. Visual tokens have their strongest impact during early generation steps and in intermediate transformer layers. Later tokens are primary text-dependent as their visual dependence decreases gradually over the generation trajectory.}
    \label{fig:teaser}
\end{figure}

In this work, we study the role of visual tokens along two axes: the generation time axis, which traces their influence over decoding steps, and the depth-wise axis, which traces the visual contribution through the transformer layers. We explore reasoning benchmarks that elicit long-form outputs, where the interaction between visual grounding and language generation can be observed over an extended trajectory. 

Our \textbf{first finding} is that the influence of visual tokens is strongest during the early stages of generation. Specifically, visual information plays a dominant role within a relatively narrow initial window, after which its influence gradually diminishes as generation progresses. In contrast, attention to previously generated text increases throughout the generation process. As a result, tokens generated in the middle and later stages rely substantially less on the original visual input and become increasingly dependent on the preceding textual context, rather than repeatedly attending to visual tokens.

Our \textbf{second finding} is that visual influence is also localized in depth during reasoning. We find visual token influence to be concentrated primarily in intermediate layers during reasoning, while its contribution becomes much less pronounced in the final layers across the entire decoding trajectory, regardless of the generation step. The final layers therefore operate predominantly on text representations, processing and selecting the next token after visual processing has already occurred in earlier visual-dependent layers. Both findings are summarized in Figure \ref{fig:teaser}. 

Overall, both our findings reveal that visual-token dependence in multimodal reasoning is neither temporally persistent nor uniformly distributed across transformer depth, and retaining the full visual context at every layer and decoder step may be unnecessary. We further verify both these findings with causal interventions, by dropping all image tokens along different times across the generation trajectory, and across different depths of the model.
\newline
\newline
In summary, our contributions are as follows: 
\begin{itemize}
    \item We analyze the visual influence in LVLMs across both time and depth in long-form reasoning benchmarks.
    \item Based on our observations, we use causal interventions to verify our findings where we drop all image tokens during different times of the multi-modal reasoning trajectory, and across different layer depths.
    \item We also provide trade-offs between visual token reduction rates and performance, across both time and depth, and across different model scales, for both image and video benchmarks.
    \item Overall, we challenge the prevailing assumption that long-form multimodal reasoning requires continuous access to the complete visual context, both in time and depth, and find that retaining all visual tokens in all layers and at every generation step is unnecessary. Our work identifies a different temporal and layer-wise pattern of visual reliance in LVLMs.
\end{itemize}

\section{Related Work}

\textbf{Visual reliance in Vision-Language Models:} Recent studies have investigated the extent to which LVLMs genuinely rely on visual information. Since the language component is typically pretrained extensively before multimodal alignment, LVLMs can exhibit strong language-driven behavior even for multimodal tasks \cite{pezeshki2021gradient,pmlr-v162-huang22e}. Several works have analyzed this phenomenon through attribution-based approaches, showing that the contribution of visual inputs to model predictions can vary substantially across tasks and examples \cite{parcalabescu2025do}. Other studies have introduced benchmarks and evaluation protocols to measure visual grounding, multimodal dependence, and susceptibility to language-only reasoning shortcuts \cite{Deng2025WordsOV,park2025generalizing,RevistingLin,Tong2024EyesWS}. However, long-form multimodal reasoning introduces a temporal dimension that is particularly important yet remains largely unexplored. Existing analyses do not examine how visual influence evolves throughout an autoregressive reasoning trajectory, whether visual information is repeatedly retrieved from the original image tokens during generation, or how visual contribution is distributed across transformer depth during reasoning. Our study aims to answer these questions while also transferring this analysis to a practical setup of interventions. 

\textbf{Visual token pruning:} A complementary line of work focuses on compressing or pruning visual tokens. Methods such as visual token pruning and selection aim to identify redundant image tokens and retain only a subset of them \cite{chen2024image,zhang2025sparsevlm,wu2026hidrop} without substantially affecting performance, particularly for short-form visual question answering (VQA) tasks. While existing pruning methods focus on identifying and removing subsets of non-informative visual tokens, our work investigates two fundamentally different questions: We first investigate whether the visual context as a whole remains necessary throughout long-form reasoning, rather than which individual visual tokens can be deemed redundant. Rather than identifying a subset of unimportant visual tokens and pruning them, we remove all visual tokens at different generation steps and transformer layers during multimodal reasoning, to determine when and where direct visual access becomes unnecessary. Secondly, prior work explores short-form VQA settings where typically a single token or brief phrase is generated and the relevant visual context is in close contextual proximity, facilitating direct access to visual evidence during generation. However, long-form multimodal reasoning introduces another temporal dimension. In this case, generation spans many decoding steps and later reasoning stages may still depend on visual evidence. Therefore, our work also studies the temporal dimension which is absent in prior works.

\begin{figure*}
    \centering
    \includegraphics[width=1\linewidth]{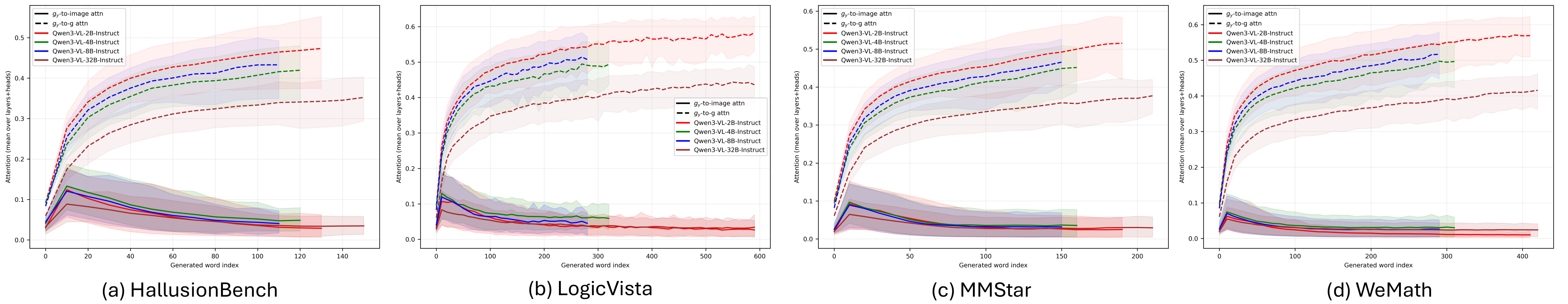}
    \caption{Attention Dynamics across Generation time using layer-head average analysis, for all models across (a) Hallusionbench, (b) LogicVista, (c) MMStar, and (d) WeMeth. For each model, we also include the variance across samples in the benchmark as a shaded region around the corresponding curve.}
    \label{fig:mean_attention}
\end{figure*}

\begin{figure*}
    \centering
    \includegraphics[width=1\linewidth]{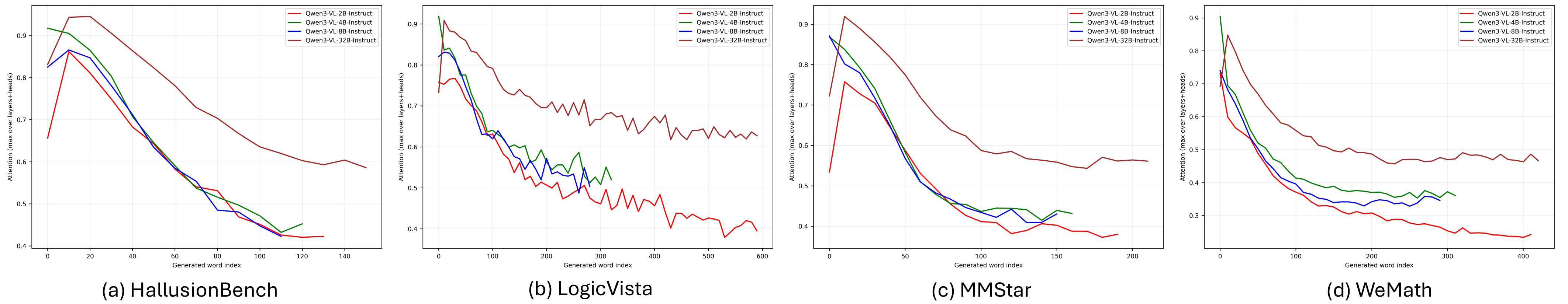}
    \caption{Attention Dynamics across Generation time using upper bound analysis, for all models across (a) Hallusionbench, (b) LogicVista, (c) MMStar, and (d) WeMeth.}
    \label{fig:max_attention}
\end{figure*}

\section{Attention Dynamics of Vision-Language Models}
\label{sec:attention_dynamics}

\textbf{Notation:} We consider an autoregressive vision-language model $M$ comprising a visual encoder $M_v$, a projection module $M_{pr}$, and a language decoder $M_{LLM}$ of $L$ layers, where $M_{LLM}$ is a large language model (LLM). We use $l$ to denote a layer and $a_l^h$ to denote the attention map of layer $l$ and head $h$. Given an input image $I$, the visual encoder divides it into $N$ visual tokens (in the form of patches) and encodes them into visual features. These features are then projected by $M_{pr}$ into the language embedding space of dimension $d$, yielding the visual representation $f_v = M_{pr}(M_v(I))$, where $f_v \in \mathbb{R}^{N \times d}$. We also denote $f_v^l$ as the visual representations at a layer $l$, and $p^l$ as the prompt representation at layer $l$. Similarly, a textual prompt $p$ comprising the problem statement and a reasoning instruction is tokenized into $T$ tokens and embedded into a textual representation $f_t \in \mathbb{R}^{T \times d}$. The visual and textual representations are then concatenated along the sequence dimension to form the input to the language decoder $M_{LLM}$, i.e., $f = [f_v; f_t]$, where $f \in \mathbb{R}^{(N+T) \times d}$. The language decoder $M_{LLM}$ autoregressively produces a reasoning response $g$ composed of $Y$ tokens, and $g_y$ denotes the $y$-th generated reasoning token. We also define $g_y^l$ as the generated token representation at a layer $l$. 
\newline
\newline
\textbf{Benchmarks:} We evaluate our approach on six benchmarks: four image-based multimodal reasoning benchmarks and two for video-based multimodal reasoning. The image benchmarks comprise Wemath \cite{wemath}, HallusionBench \cite{hallusionbench}, MMStar \cite{mmstar} and LogicVista \cite{logicvista}. WeMath assesses human-like mathematical reasoning through hierarchical visual-mathematics problems. HallusionBench evaluates reasoning robustness against entangled visual illusions and language hallucinations. MMStar measures multimodal capabilities across a balanced set of perception and reasoning-oriented tasks. LogicVista focuses on logical reasoning grounded in diverse visual contexts. For video benchmarks, we use MMVU \cite{Zhao2025MMVUME} and MVBench \cite{Li2023MVBenchAC}.  MMVU evaluates expert-level video understanding and domain-specific reasoning across disciplines such as science, healthcare, and engineering. MVBench assesses temporal reasoning through 20 video understanding tasks (e.g., motion understanding, event sequencing). For each video, we sample one frame per second. 
\newline
\newline
\textbf{Models:} We conduct experiments using 8 LVLMs in total. We first consider the Qwen3-VL family \cite{Qwen3-VL} with scales 2B, 4B, 8B and 32B. We also consider the InternVL-3.5 family of models \cite{Wang2025InternVL35AO} with scales 4B, 8B and 38B. Finally, we also consider the Molmo2-8B \cite{Clark2026Molmo2OW} model. We detail the prompts we used in Appendix \ref{app:prompts}. 

\subsection{Attention Dynamics across Generation Time}
\label{subsection:dynamics_generation_time}

\begin{figure*}
    \centering
    \includegraphics[width=1\linewidth]{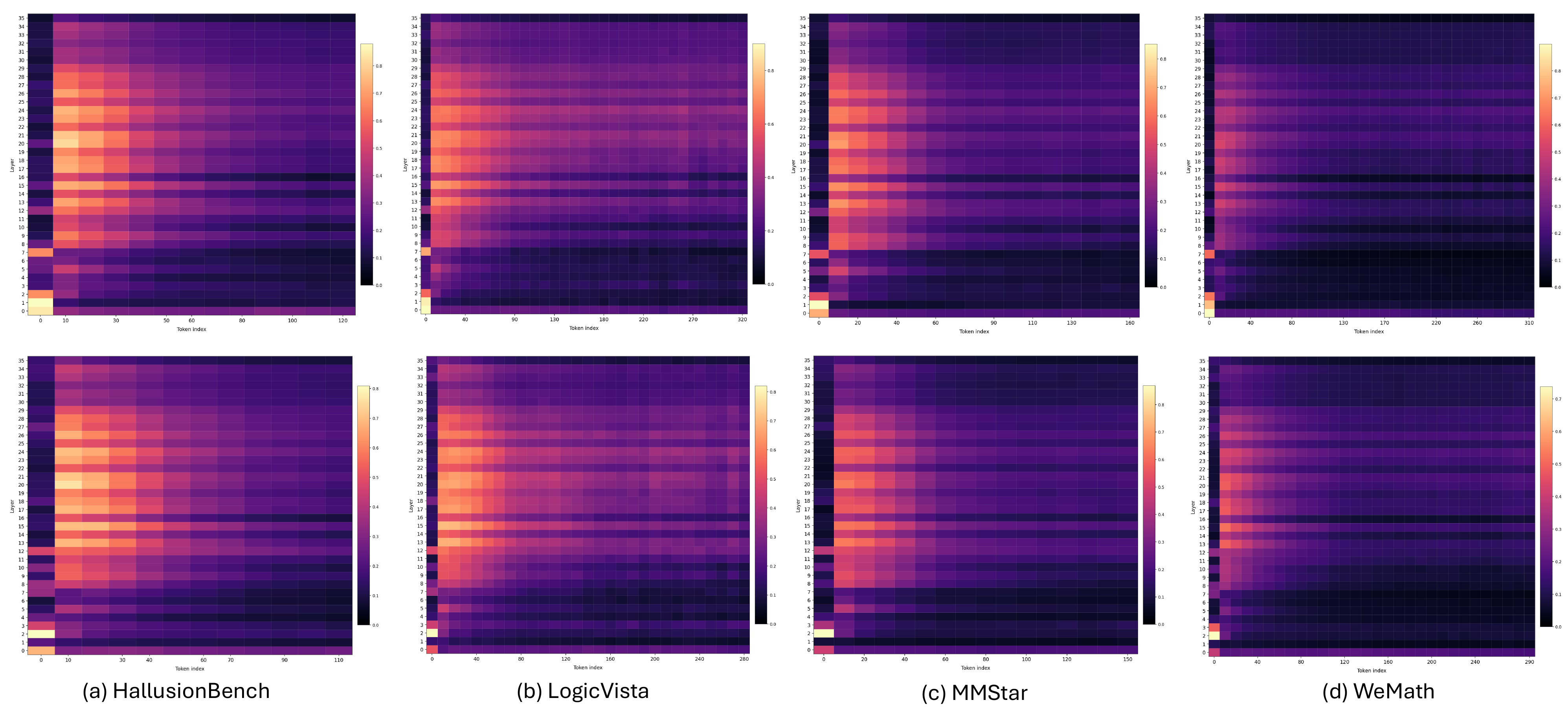}
    \caption{Attention Dynamics across Transformer Depth for Qwen3-VL-4B for (a) Hallusionbench, (b) LogicVista, (c) MMStar, and (d) WeMeth.}
    \label{fig:4B_8B_layerwise_heatmaps}
\end{figure*}

\begin{figure}
    \centering
    \includegraphics[width=1\linewidth]{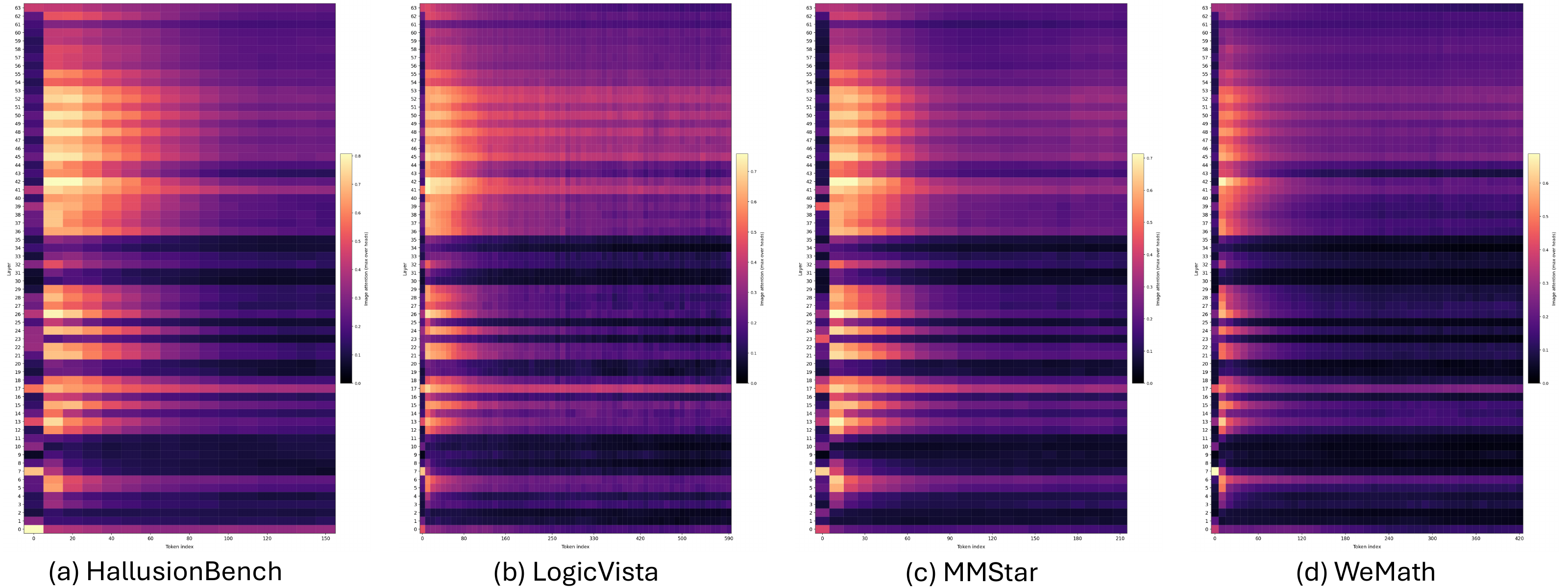}
    \caption{Attention Dynamics across Transformer Depth for Qwen3-VL-32B for (a) Hallusionbench, (b) LogicVista, (c) MMStar, and (d) WeMeth.}
    \label{fig:32B_layerwise_heatmaps}
\end{figure}

We analyze the attention dynamics of the generated reasoning chain $g$ across generation time. For each generated reasoning token $g_y$, we examine its interaction with (i) the image tokens $f_v$ and (ii) the previously generated reasoning tokens $g_{0 \dots y-1}$, which we denote as $g_{<y}$. Concretely, at a layer $l$ and an attention head $h$, the attention map $a_l^h$ for token $g_y$ distributes attention across the $N$ image tokens ($f_v^l$), the $T$ text prompt tokens ($p^l$), and the preceding generated tokens $g_{<y}^l$. We denote $a_l^h(S) := \{a_l^h[i]\}_{i \in S}$ as the attention placed on a token subset $S$. We decompose $a_l^h$ for $g_y$ into three components (three subsets): (1) attention allocation to the visuial tokens, $a^{img} = a_l^h(f_v^l)$, (2) attention to the text prompt, $a^{pmt} = a_l^h(p^l)$, and (3) attention to previously generated reasoning tokens, $a^{gen} = a_l^h(g_{<y}^l)$, where $y$ is the index of the current generated token. Since the generated reasoning sequences may differ in length across samples, we first synchronize them according to their token positions using padding. Statistical quantities are then computed only over non-padded entries, ensuring that each estimate reflects contributions from valid tokens only. Finally, to avoid unreliable behavior in the long-tail region, we restrict the analysis to token positions supported by a sufficient number of samples. Specifically, a token index is included only if it is present in at least 20\% of the samples. As a result, different curves may end at different token positions.

Figure~\ref{fig:mean_attention} plots $a^{img}$ and $a^{gen}$ attention scores as functions of the generated reasoning tokens for all four benchmarks across different models, averaging over all layers and heads and then averaging across all samples in the corresponding dataset. Because the attention weights sum to one, the attention from $g_y$ to the $T$ text tokens in the prompt $p$ can be recovered by $a^{pmt}=1-\left(\sum_{f_v}a^{img}+\sum_{g_{<y}}a^{gen}\right)$; because our focus is the contribution of visual attention as tokens are generated, we omit this term from the plots.

Across all four benchmarks and models, attention to the image tokens shows a consistent early peak, reaching its maximum within roughly the first few tens of generated tokens before steadily decreasing as generation proceeds, whereas attention to previously generated tokens rises sharply during the same initial period and continues increasing throughout the response, becoming dominant during the middle and late reasoning steps. The curves suggest a relatively short window (Figure \ref{fig:teaser}), concentrated early within the first few tens of generated tokens, where the generated tokens attend strongly to the visual tokens and visual dependence is strong. After this window, the model appears to rely increasingly on previously generated textual tokens, while direct attention to the original visual tokens diminishes. This behavior indicates that the visual tokens become progressively redundant as decoding proceeds, suggesting that maintaining full access to all visual tokens throughout the entire generation trajectory may be unnecessary after the early window. 

In order to obtain an \textit{upper bound} of the visual token influence on the generated tokens, for each generated token $g_y$, we take the maximum value across all layers and heads and average those maximum values across all samples in the dataset. Results are shown in Figure \ref{fig:max_attention}. Even with the upper bound, the visual token influence gradually drops across all models and benchmarks, further supporting our finding. Interestingly, we also observe that larger models tend to exhibit higher maximum visual token influence than smaller models, suggesting that larger models rely more heavily on visual evidence.

\begin{tcolorbox}[
    colback=green!8,
    colframe=green!50!black,
    boxrule=0.8pt,
    arc=2mm,
    left=2mm,
    right=2mm,
    top=1mm,
    bottom=1mm
]
\textbf{Observation 1:} Visual influence peaks early during the first few tens of generated tokens and gradually decreases, after which reasoning becomes increasingly dominated by previously generated textual context.
\end{tcolorbox}

\begin{figure*}
    \centering
    \includegraphics[width=1\linewidth]{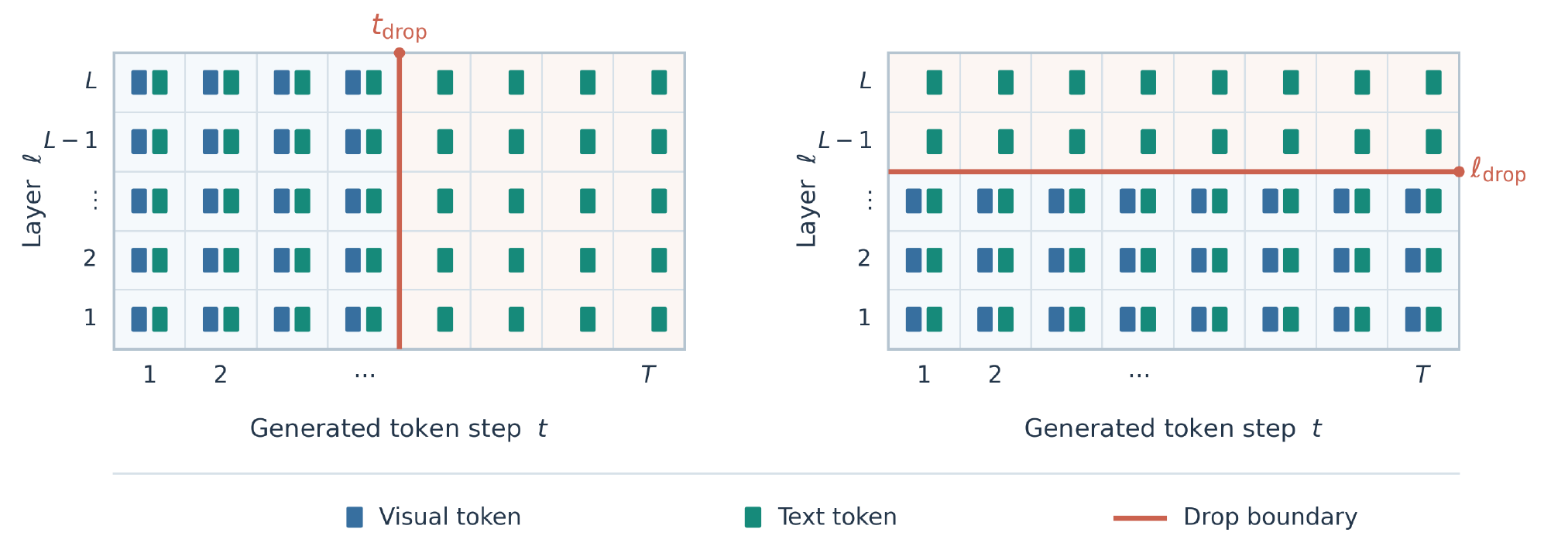}
    \caption{Causal Interventions through visual token removal across generation time (left) and transformer depth (right).}
    \label{fig:method}
\end{figure*}

\subsection{Attention Dynamics across Transformer Depth}
\label{subsec:attention_dynamics_across_transformer_depth}

Next, we analyze the attention dynamics of the generated reasoning chain $g$ across the $L$ transformer layers. For each generated reasoning token $g_y$ and each layer $l$, we select the maximum attention from $g_y$ to the visual tokens across all attention heads, and then average across all samples in the corresponding dataset. The results are shown in Figure \ref{fig:4B_8B_layerwise_heatmaps} for Qwen3-VL-4B and Qwen3-VL-8B, respectively. Both Qwen3-VL-4B and Qwen3-VL-8B consist of 36 transformer layers, shown on the y-axis. The x-axis represents generation time (i.e., the position of each generated reasoning token in the sequence). The heatmap encodes the strength of maximum attention to the visual tokens, with brighter regions indicating stronger visual attention.

Across all benchmarks and models, we observe a consistent pattern. Visual influence is concentrated primarily in the intermediate transformer layers, whereas both the early and late layers exhibit substantially weaker attention to the visual tokens. At the same time, we observe a clear temporal trend across the reasoning trajectory, mirroring the findings in Section \ref{subsection:dynamics_generation_time}. Visual attention is strongest for the first few generated reasoning tokens and gradually decreases as generation progresses. The last layers, have a relatively weaker visual influence across the entire decoding trajectory. These results suggest that the high visual dependence is localized not only in generation time but also in transformer depth, with visual dependence being strong primarily in intermediate layers during the early stages of decoding, while the final layers operate predominantly on the previous textual representations. 

To investigate whether this phenomenon generalizes to larger models, we conduct the same analysis on Qwen3-VL-32B. As shown in Figure \ref{fig:32B_layerwise_heatmaps}, we observe the same pattern, indicating that the same layer-wise and temporal dynamics persist at a larger model scale.

\begin{tcolorbox}[
    colback=green!8,
    colframe=green!50!black,
    boxrule=0.8pt,
    arc=2mm,
    left=2mm,
    right=2mm,
    top=1mm,
    bottom=1mm
]
\textbf{Observation 2:} Visual influence peaks in intermediate transformer layers during early reasoning, and progressively weakens across both generation time and later layers.
\end{tcolorbox}

\section{Causal Interventions}
\label{sec:causual_intervention}

\begin{table*}[t]
\centering
\caption{Visual token dropping at different generation steps across four benchmarks and eight vision-language models.
We indicate the \colorbox{red!28.8}{degradation} from the baseline, where darker red indicates higher degradation. The \textcolor{green!50!black}{prompt token reduction (\%)} is shown next to each score.}
\label{tab:dropping-generation-time}

\begingroup
\setlength{\tabcolsep}{3pt}
\setlength{\fboxsep}{1pt}

\begin{adjustbox}{max width=\textwidth}
\begin{tabular}{llcccccccc}
\toprule
\textbf{Benchmark} & \textbf{Drop}
& \multicolumn{4}{c}{\textbf{Qwen3-VL}}
& \multicolumn{1}{c}{\textbf{Molmo-2}}
& \multicolumn{3}{c}{\textbf{InternVL-3.5}} \\
\cmidrule(lr){3-6}
\cmidrule(lr){7-7}
\cmidrule(lr){8-10}
& & \textbf{2B} & \textbf{4B} & \textbf{8B} & \textbf{32B}
& \textbf{8B}
& \textbf{4B} & \textbf{8B} & \textbf{38B} \\
\midrule


\textbf{WeMath}
& Baseline
& 57.13\nodrop
& 68.85\nodrop
& 72.30\nodrop
& 74.31\nodrop
& 56.32\nodrop
& 67.76\nodrop
& 72.01\nodrop
& 74.83\nodrop \\ \cline{2-10}

& 50
& \colorbox{red!28.8}{48.33}\drop{31.6}
& \colorbox{red!28.5}{60.23}\drop{31.3}
& \colorbox{red!26.8}{64.60}\drop{32.0}
& \colorbox{red!24.8}{67.64}\drop{29.8}
& \colorbox{red!15.8}{53.10}\drop{55.21}
& \colorbox{red!60.0}{59.89}\drop{53.77}
& \colorbox{red!57.4}{64.83}\drop{52.01}
& \colorbox{red!29.4}{71.15}\drop{57.55} \\

& 100
& \colorbox{red!18.3}{53.28}\drop{33.5}
& \colorbox{red!18.6}{64.89}\drop{30.6}
& \colorbox{red!13.8}{69.89}\drop{28.3}
& \colorbox{red!18.8}{70.29}\drop{32.9}
& \colorbox{red!1.8}{56.09}\drop{28.71}
& \colorbox{red!22.6}{64.94}\drop{51.54}
& \colorbox{red!24.3}{68.97}\drop{49.41}
& \colorbox{red!7.4}{73.91}\drop{46.64} \\

& 200
& \colorbox{red!7.3}{56.09}\drop{17.7}
& \colorbox{red!6.0}{67.99}\drop{11.6}
& \colorbox{red!3.6}{71.67}\drop{11.1}
& \colorbox{red!4.4}{73.62}\drop{17.3}
& \colorbox{red!0.5}{56.26}\drop{5.45}
& \colorbox{red!6.0}{67.01}\drop{29.83}
& \colorbox{red!10.1}{70.75}\drop{30.23}
& \colorbox{red!2.8}{74.48}\drop{22.41} \\

& 300
& 57.07\drop{10.9}
& 68.39\drop{7.2}
& 71.95\drop{6.9}
& 73.91\drop{10.0}
& \colorbox{red!0.5}{56.26}\drop{1.81}
& \colorbox{red!3.7}{67.30}\drop{15.41}
& \colorbox{red!2.7}{71.67}\drop{15.03}
& \colorbox{red!2.3}{74.54}\drop{13.81} \\

\midrule


\textbf{LogicVista}
& Baseline
& 33.71\nodrop
& 39.73\nodrop
& 50.89\nodrop
& 51.56\nodrop
& 41.29\nodrop
& 46.65\nodrop
& 46.21\nodrop
& 55.13\nodrop \\ \cline{2-10}

& 50
& \colorbox{red!30.2}{24.11}\drop{44.4}
& \colorbox{red!35.9}{26.34}\drop{41.2}
& \colorbox{red!40.0}{34.38}\drop{39.7}
& \colorbox{red!48.1}{27.90}\drop{41.9}
& \colorbox{red!34.4}{36.61}\drop{59.42}
& \colorbox{red!60.0}{37.05}\drop{53.92}
& \colorbox{red!60.0}{37.05}\drop{57.67}
& \colorbox{red!60.0}{43.53}\drop{59.50} \\

& 100
& \colorbox{red!17.5}{30.13}\drop{48.0}
& \colorbox{red!30.9}{29.69}\drop{43.4}
& \colorbox{red!27.4}{42.86}\drop{41.1}
& \colorbox{red!39.5}{35.49}\drop{40.3}
& \colorbox{red!12.0}{39.29}\drop{19.77}
& \colorbox{red!30.3}{42.86}\drop{49.80}
& \colorbox{red!14.3}{44.42}\drop{49.18}
& \colorbox{red!50.0}{48.88}\drop{43.96} \\

& 200
& \colorbox{red!10.3}{32.14}\drop{38.8}
& \colorbox{red!7.8}{38.62}\drop{23.2}
& \colorbox{red!11.3}{49.11}\drop{18.2}
& \colorbox{red!22.5}{45.98}\drop{37.5}
& 41.29\drop{2.57}
& \colorbox{red!8.9}{45.54}\drop{25.47}
& \colorbox{red!9.0}{45.09}\drop{18.09}
& \colorbox{red!8.9}{54.02}\drop{17.40} \\

& 300
& 33.48\drop{26.5}
& \colorbox{red!4.1}{39.06}\drop{12.8}
& 50.67\drop{9.6}
& \colorbox{red!18.8}{47.54}\drop{28.2}
& 41.29\drop{0.35}
& \colorbox{red!5.4}{45.98}\drop{10.19}
& 46.43\drop{5.20}
& \colorbox{red!5.4}{54.46}\drop{6.52} \\

\midrule


\textbf{MMStar}
& Baseline
& 56.13\nodrop
& 66.67\nodrop
& 68.00\nodrop
& 62.33\nodrop
& 64.00\nodrop
& 66.87\nodrop
& 66.87\nodrop
& 72.87\nodrop \\ \cline{2-10}

& 50
& \colorbox{red!17.8}{52.47}\drop{38.7}
& \colorbox{red!19.2}{62.47}\drop{38.6}
& \colorbox{red!18.2}{64.20}\drop{39.0}
& \colorbox{red!22.3}{56.87}\drop{35.9}
& \colorbox{red!15.8}{61.67}\drop{53.43}
& \colorbox{red!20.8}{64.27}\drop{61.60}
& \colorbox{red!22.4}{64.07}\drop{50.37}
& \colorbox{red!8.0}{71.87}\drop{57.64} \\

& 100
& \colorbox{red!6.6}{55.20}\drop{31.7}
& \colorbox{red!9.8}{65.20}\drop{24.3}
& \colorbox{red!10.8}{66.33}\drop{22.4}
& \colorbox{red!10.1}{60.80}\drop{35.2}
& \colorbox{red!1.6}{63.80}\drop{8.84}
& \colorbox{red!3.2}{66.47}\drop{23.58}
& \colorbox{red!4.8}{66.27}\drop{14.97}
& \colorbox{red!2.7}{72.53}\drop{20.29} \\

& 200
& 56.27\drop{9.8}
& \colorbox{red!4.9}{65.93}\drop{6.3}
& 68.00\drop{5.5}
& \colorbox{red!8.4}{61.13}\drop{8.3}
& 64.00\drop{1.93}
& 67.00\drop{7.03}
& \colorbox{red!1.1}{66.73}\drop{6.34}
& \colorbox{red!2.2}{72.60}\drop{7.61} \\

& 300
& 56.13\drop{8.0}
& 66.47\drop{4.7}
& 67.60\drop{3.9}
& \colorbox{red!1.7}{61.80}\drop{4.0}
& 64.07\drop{1.22}
& 67.07\drop{4.16}
& 67.20\drop{3.39}
& 72.87\drop{5.33} \\

\midrule


\textbf{HallusionBench}
& Baseline
& 67.82\nodrop
& 74.55\nodrop
& 73.82\nodrop
& 75.71\nodrop
& 64.88\nodrop
& 70.56\nodrop
& 70.56\nodrop
& 74.24\nodrop \\ \cline{2-10}

& 50
& \colorbox{red!7.4}{66.77}\drop{59.0}
& \colorbox{red!11.8}{72.66}\drop{54.3}
& 73.92\drop{53.5}
& \colorbox{red!9.8}{74.24}\drop{53.1}
& \colorbox{red!1.7}{64.67}\drop{48.30}
& \colorbox{red!5.9}{69.82}\drop{66.12}
& \colorbox{red!9.3}{69.40}\drop{54.51}
& \colorbox{red!3.4}{73.82}\drop{55.21} \\

& 100
& 67.40\drop{30.9}
& 74.66\drop{26.6}
& 73.61\drop{22.8}
& 75.50\drop{44.4}
& 64.88\drop{3.31}
& \colorbox{red!3.4}{70.14}\drop{17.34}
& \colorbox{red!1.7}{70.35}\drop{11.67}
& 74.34\drop{12.15} \\

& 200
& 68.14\drop{4.8}
& 74.24\drop{1.3}
& 73.92\drop{1.3}
& 75.60\drop{4.6}
& 64.88\drop{0.17}
& 70.56\drop{1.17}
& 70.56\drop{0.00}
& 74.34\drop{2.45} \\

& 300
& 67.82\drop{3.2}
& 74.55\drop{0.6}
& 73.82\drop{0.1}
& 75.60\drop{1.3}
& 64.88\drop{0.17}
& 70.56\drop{0.00}
& 70.56\drop{0.00}
& 74.24\drop{1.40} \\

\bottomrule
\end{tabular}
\end{adjustbox}

\endgroup
\end{table*}

\begin{figure*}
    \centering
    \includegraphics[width=1\linewidth]{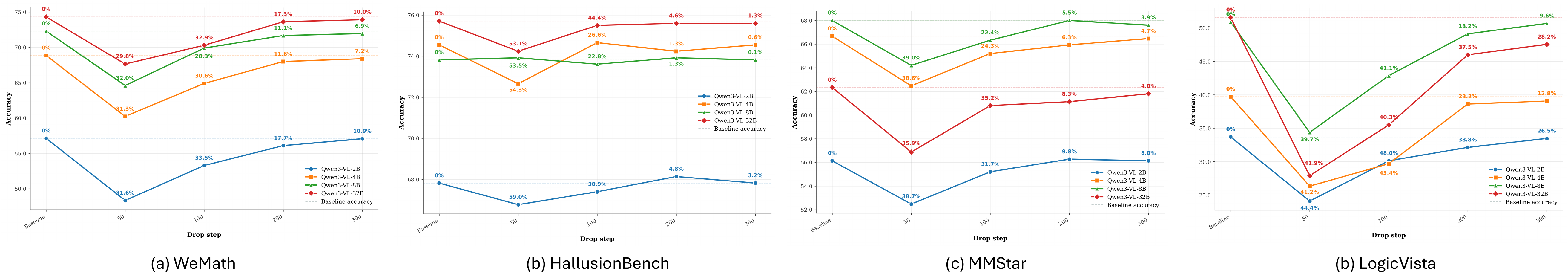}
    \caption{Visual token dropping across generation time for (a) Wemath, (b) HallusionBench. Different models are shown in different colors. The x-axis denotes different dropping times $t_{drop}$ while the y-axis shows the accuracy. The prompt token reduction \% is shown above each point.}
    \label{fig:img_kv_cache_drop}
\end{figure*}

\begin{table*}[t]
\centering
\caption{Visual token dropping at different layers across four benchmarks
and eight models. We drop at layer $L$ - offset, where $L$ is the final
layer index. Red shading indicates \colorbox{red!20}{degradation}
from the baseline, with darker red indicating higher degradation.
The \textcolor{green!50!black}{all-layer visual token reduction \%}
is shown next to each score.}
\label{tab:image_token_drop}

\begin{adjustbox}{max width=\textwidth}
\begin{tabular}{llcccccccc}
\toprule

& &
\multicolumn{4}{c}{\textbf{Qwen3-VL}} &
\multicolumn{1}{c}{\textbf{Molmo-2}} &
\multicolumn{3}{c}{\textbf{InternVL-3.5}} \\

\cmidrule(lr){3-6}
\cmidrule(lr){7-7}
\cmidrule(lr){8-10}

\textbf{Benchmark} & \textbf{Drop}
& \textbf{2B}
& \textbf{4B}
& \textbf{8B}
& \textbf{32B}
& \textbf{8B}
& \textbf{4B}
& \textbf{8B}
& \textbf{38B} \\

\midrule


\textbf{WeMath}
& Baseline
& 57.13\nodrop
& 68.85\nodrop
& 72.30\nodrop
& 74.31\nodrop
& 56.32\nodrop
& 67.76\nodrop
& 72.01\nodrop
& 74.83\nodrop \\

\cline{2-10}

& Offset 5
& \colorbox{red!10.8}{55.46}\drop{11.71}
& \colorbox{red!2.6}{68.28}\drop{9.10}
& 73.05\drop{9.77}
& 74.54\drop{5.36}
& 56.38\drop{9.21}
& 68.22\drop{9.31}
& \colorbox{red!5.04}{71.38}\drop{9.47}
& 75.23\drop{5.91} \\

& Offset 10
& \colorbox{red!41.7}{39.25}\drop{19.11}
& \colorbox{red!19.8}{64.43}\drop{16.43}
& \colorbox{red!11.3}{70.52}\drop{16.22}
& 74.25\drop{10.06}
& \colorbox{red!37.68}{51.61}\drop{16.08}
& \colorbox{red!18.40}{65.46}\drop{16.78}
& \colorbox{red!17.92}{69.77}\drop{16.68}
& \colorbox{red!5.52}{74.14}\drop{10.77} \\

& Offset 15
& \colorbox{red!50.0}{30.29}\drop{28.23}
& \colorbox{red!46.4}{46.78}\drop{18.58}
& \colorbox{red!44.4}{52.13}\drop{19.80}
& \colorbox{red!44.6}{53.91}\drop{11.05}
& \colorbox{red!79.52}{46.38}\drop{21.79}
& \colorbox{red!80.00}{54.60}\drop{21.55}
& \colorbox{red!80.00}{59.60}\drop{20.64}
& \colorbox{red!72.24}{65.80}\drop{13.54} \\

\midrule


\textbf{LogicVista}
& Baseline
& 33.71\nodrop
& 39.73\nodrop
& 50.89\nodrop
& 51.56\nodrop
& 41.29\nodrop
& 46.65\nodrop
& 46.21\nodrop
& 55.13\nodrop \\

\cline{2-10}

& Offset 5
& \colorbox{red!14.0}{31.25}\drop{11.52}
& 40.40\drop{8.56}
& \colorbox{red!16.9}{47.54}\drop{8.46}
& 51.12\drop{4.67}
& 42.41\drop{7.47}
& 47.77\drop{7.50}
& 46.43\drop{7.87}
& \colorbox{red!7.12}{54.24}\drop{4.95} \\

& Offset 10
& \colorbox{red!34.0}{21.65}\drop{19.05}
& \colorbox{red!25.8}{32.59}\drop{15.56}
& \colorbox{red!32.3}{39.96}\drop{15.07}
& \colorbox{red!14.0}{49.11}\drop{9.07}
& \colorbox{red!74.96}{31.92}\drop{15.69}
& \colorbox{red!48.24}{40.62}\drop{15.75}
& \colorbox{red!35.76}{41.74}\drop{14.57}
& 55.36\drop{9.10} \\

& Offset 15
& \colorbox{red!27.9}{25.45}\drop{39.58}
& \colorbox{red!42.7}{20.98}\drop{20.28}
& \colorbox{red!49.3}{26.12}\drop{19.62}
& \colorbox{red!50.0}{25.67}\drop{9.94}
& \colorbox{red!80.00}{27.68}\drop{23.25}
& \colorbox{red!80.00}{36.61}\drop{22.14}
& \colorbox{red!80.00}{33.26}\drop{23.07}
& \colorbox{red!80.00}{43.75}\drop{13.13} \\

\midrule


\textbf{MMStar}
& Baseline
& 56.13\nodrop
& 66.67\nodrop
& 68.00\nodrop
& 62.33\nodrop
& 64.00\nodrop
& 66.87\nodrop
& 66.87\nodrop
& 72.87\nodrop \\

\cline{2-10}

& Offset 5
& 55.87\drop{10.82}
& \colorbox{red!6.1}{65.80}\drop{8.29}
& \colorbox{red!1.7}{67.47}\drop{8.96}
& \colorbox{red!16.6}{59.07}\drop{5.04}
& \colorbox{red!12.24}{62.47}\drop{8.85}
& 67.13\drop{8.50}
& \colorbox{red!8.56}{65.80}\drop{9.39}
& \colorbox{red!1.60}{72.67}\drop{5.59} \\

& Offset 10
& \colorbox{red!31.1}{45.93}\drop{19.67}
& \colorbox{red!19.1}{62.53}\drop{15.72}
& \colorbox{red!17.4}{64.47}\drop{16.44}
& \colorbox{red!13.0}{60.13}\drop{9.28}
& \colorbox{red!17.04}{61.87}\drop{16.78}
& \colorbox{red!20.80}{64.27}\drop{16.11}
& \colorbox{red!15.52}{64.93}\drop{17.11}
& \colorbox{red!3.20}{72.47}\drop{10.26} \\

& Offset 15
& \colorbox{red!37.4}{41.67}\drop{31.75}
& \colorbox{red!32.6}{55.53}\drop{22.58}
& \colorbox{red!32.1}{57.20}\drop{23.81}
& \colorbox{red!33.9}{50.33}\drop{12.70}
& \colorbox{red!77.84}{54.27}\drop{25.19}
& \colorbox{red!41.60}{61.67}\drop{24.14}
& \colorbox{red!37.92}{62.13}\drop{25.42}
& \colorbox{red!20.32}{70.33}\drop{13.81} \\

\midrule


\textbf{HallusionBench}
& Baseline
& 67.82\nodrop
& 74.55\nodrop
& 73.82\nodrop
& 75.71\nodrop
& 64.88\nodrop
& 70.56\nodrop
& 70.56\nodrop
& 74.24\nodrop \\

\cline{2-10}

& Offset 5
& 68.45\drop{10.76}
& \colorbox{red!11.3}{72.77}\drop{7.97}
& 73.50\drop{8.56}
& 75.71\drop{4.87}
& \colorbox{red!17.68}{62.67}\drop{9.20}
& \colorbox{red!2.56}{70.24}\drop{9.19}
& \colorbox{red!9.28}{69.40}\drop{9.48}
& \colorbox{red!5.92}{73.50}\drop{5.73} \\

& Offset 10
& \colorbox{red!22.7}{62.15}\drop{19.60}
& \colorbox{red!21.6}{69.40}\drop{15.01}
& \colorbox{red!14.2}{71.29}\drop{15.36}
& \colorbox{red!12.6}{73.61}\drop{9.05}
& \colorbox{red!4.24}{64.35}\drop{16.53}
& \colorbox{red!28.64}{66.98}\drop{17.91}
& \colorbox{red!15.20}{68.66}\drop{17.32}
& \colorbox{red!8.40}{73.19}\drop{10.61} \\

& Offset 15
& \colorbox{red!35.3}{54.89}\drop{33.86}
& \colorbox{red!24.5}{68.03}\drop{21.12}
& \colorbox{red!21.1}{68.87}\drop{22.65}
& \colorbox{red!24.3}{69.30}\drop{13.16}
& \colorbox{red!37.04}{60.25}\drop{24.93}
& \colorbox{red!32.80}{66.46}\drop{28.08}
& \colorbox{red!27.76}{67.09}\drop{26.95}
& \colorbox{red!37.84}{69.51}\drop{14.29} \\

\bottomrule

\end{tabular}
\end{adjustbox}
\end{table*}

\begin{figure*}
    \centering
    \includegraphics[width=1\linewidth]{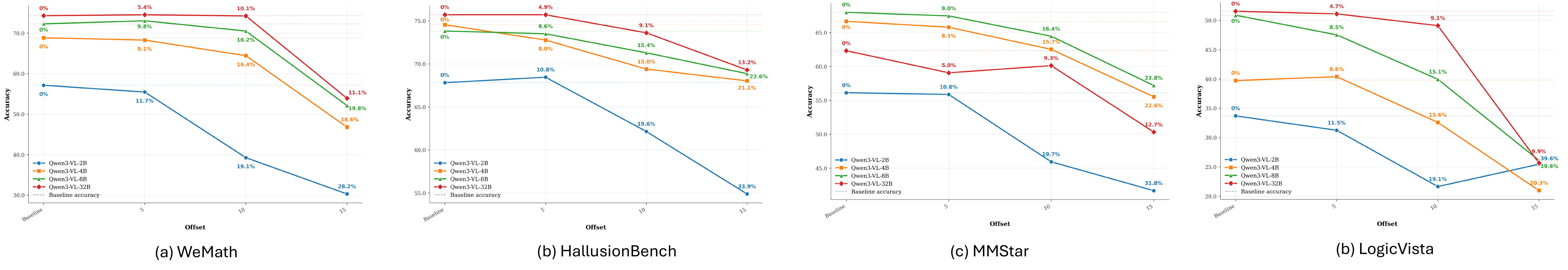}
    \caption{Visual token dropping across transformer depth for (a) WeMath, (b) HallusionBench. The x-axis denotes different layer offsets, where visual tokens are removed starting from the last \(L-\mathrm{offset}\) layers. The y-axis shows the accuracy. The all-layer visual token reduction \% is shown next to each point.}
    \label{fig:img_token_drop}
\end{figure*}

\begin{table*}[t]
\centering
\caption{Video token dropping at different generation steps
across two benchmarks and four Qwen3-VL models.
For each model, the first column reports accuracy with
\textcolor{green!50!black}{prompt token reduction (\%)},
while the second reports
\textcolor{blue!70!black}{KV-cache reduction (\%)}
and \textcolor{brown!80!black}{maximum memory savings (GiB)}
separated by a slash.
Memory savings are reported using $1024^3$ bytes per GiB.
Red shading indicates accuracy degradation relative to
the baseline, with darker red indicating greater degradation.}

\begingroup
\setlength{\tabcolsep}{3pt}
\renewcommand{\arraystretch}{1.12}
\small

\begin{adjustbox}{max width=\textwidth}
\begin{tabular}{ll*{4}{cc}}
\toprule
\textbf{Benchmark} & \textbf{Drop}
& \multicolumn{8}{c}{\textbf{Qwen3-VL}} \\
\cmidrule(lr){3-10}

& & \multicolumn{2}{c}{\textbf{2B}}
& \multicolumn{2}{c}{\textbf{4B}}
& \multicolumn{2}{c}{\textbf{8B}}
& \multicolumn{2}{c}{\textbf{32B}} \\

\cmidrule(lr){3-4}
\cmidrule(lr){5-6}
\cmidrule(lr){7-8}
\cmidrule(lr){9-10}

& & \textbf{Acc. (\%)} & \textbf{KV\% / GiB}
  & \textbf{Acc. (\%)} & \textbf{KV\% / GiB}
  & \textbf{Acc. (\%)} & \textbf{KV\% / GiB}
  & \textbf{Acc. (\%)} & \textbf{KV\% / GiB} \\

\midrule

\textbf{MMVU} & Baseline
& 39.94\nodrop & \kvmem{0.00}{0.000}
& 52.03\nodrop & \kvmem{0.00}{0.000}
& 56.64\nodrop & \kvmem{0.00}{0.000}
& 64.45\nodrop & \kvmem{0.00}{0.000} \\
\cline{2-10}

& 25
& 41.86\drop{58.5} & \kvmem{93.29}{1.304}
& \colorbox{red!11.2}{49.79}\drop{57.8} & \kvmem{93.29}{1.677}
& \colorbox{red!7.5}{55.14}\drop{55.9} & \kvmem{93.29}{1.677}
& \colorbox{red!15.0}{61.46}\drop{58.1} & \kvmem{93.29}{2.981} \\

& 50
& 40.04\drop{65.0} & \kvmem{92.81}{1.304}
& \colorbox{red!1.6}{51.71}\drop{59.4} & \kvmem{92.81}{1.677}
& \colorbox{red!3.2}{56.00}\drop{58.8} & \kvmem{92.81}{1.677}
& \colorbox{red!4.8}{63.49}\drop{57.0} & \kvmem{92.81}{2.981} \\

& 100
& 40.36\drop{49.1} & \kvmem{91.54}{1.304}
& 52.14\drop{50.9} & \kvmem{91.75}{1.677}
& 56.85\drop{43.8} & \kvmem{91.58}{1.677}
& \colorbox{red!1.6}{64.13}\drop{59.9} & \kvmem{91.63}{2.981} \\

& 200
& 39.94\drop{12.9} & \kvmem{90.40}{1.264}
& 52.03\drop{2.8} & \kvmem{89.24}{1.585}
& 56.75\drop{4.1} & \kvmem{90.59}{1.625}
& \colorbox{red!0.5}{64.35}\drop{8.2} & \kvmem{89.57}{2.844} \\

& 300
& 39.94\drop{9.9} & \kvmem{89.24}{1.264}
& 52.03\drop{1.5} & \kvmem{87.28}{1.372}
& 56.64\drop{1.0} & \kvmem{88.47}{1.402}
& 64.45\drop{2.3} & \kvmem{87.35}{2.710} \\

\midrule

\textbf{MVBench} & Baseline
& 39.50\nodrop & \kvmem{0.00}{0.000}
& 52.50\nodrop & \kvmem{0.00}{0.000}
& 50.50\nodrop & \kvmem{0.00}{0.000}
& 68.00\nodrop & \kvmem{0.00}{0.000} \\
\cline{2-10}

& 25
& \colorbox{red!2.5}{39.00}\drop{59.0} & \kvmem{98.18}{1.365}
& \colorbox{red!15.0}{49.50}\drop{57.2} & \kvmem{98.18}{1.755}
& \colorbox{red!7.5}{49.00}\drop{58.3} & \kvmem{98.18}{1.755}
& \colorbox{red!37.5}{60.50}\drop{45.5} & \kvmem{98.18}{3.121} \\

& 50
& 40.00\drop{66.2} & \kvmem{97.98}{1.365}
& \colorbox{red!20.0}{48.50}\drop{70.1} & \kvmem{97.95}{1.755}
& 52.50\drop{65.2} & \kvmem{97.95}{1.755}
& \colorbox{red!27.5}{62.50}\drop{51.6} & \kvmem{97.96}{3.121} \\

& 100
& \colorbox{red!5.0}{38.50}\drop{30.0} & \kvmem{97.57}{1.365}
& \colorbox{red!2.5}{52.00}\drop{15.5} & \kvmem{97.42}{1.755}
& 51.00\drop{14.8} & \kvmem{97.50}{1.755}
& \colorbox{red!15.0}{65.00}\drop{38.7} & \kvmem{97.45}{3.121} \\

& 200
& 39.50\drop{18.2} & \kvmem{96.77}{1.365}
& 52.50\drop{6.8} & \kvmem{96.37}{1.712}
& \colorbox{red!2.5}{50.00}\drop{4.4} & \kvmem{96.76}{1.755}
& \colorbox{red!2.5}{67.50}\drop{4.8} & \kvmem{96.00}{3.043} \\

& 300
& 39.50\drop{18.2} & \kvmem{95.94}{1.365}
& 52.50\drop{5.8} & \kvmem{95.36}{1.712}
& 50.50\drop{2.0} & \kvmem{96.03}{1.712}
& 68.00\drop{1.5} & \kvmem{95.58}{2.997} \\

\bottomrule
\end{tabular}
\end{adjustbox}

\endgroup
\label{tab:dropping-generation-time-video-benchmarks}
\end{table*}

\begin{table*}
\setlength{\tabcolsep}{3pt}
\renewcommand{\arraystretch}{1.12}
\small
\centering
\caption{Visual token dropping at different layers across two benchmarks and four models. Dropping starts at layer $L-\mathrm{offset}$, where $L$ is the final layer index. For each model, the first column reports accuracy with
\textcolor{green!50!black}{all-layer visual token reduction \%},
while the second reports \textcolor{blue!70!black}{KV-cache reduction (\%)} and \textcolor{brown!80!black}{maximum memory savings (GiB)}
separated by a slash. Memory savings are reported using $1024^3$ bytes per GiB. Red shading indicates accuracy degradation relative to
the baseline, with darker red indicating greater degradation.
}
\label{tab:image_token_drop_video}
\begin{adjustbox}{max width=\textwidth}
\begin{tabular}{ll*{4}{cc}}
\toprule
\multirow{3}{*}{\textbf{Benchmark}} & \multirow{3}{*}{\textbf{Drop}} & \multicolumn{8}{c}{\textbf{Qwen3-VL}} \\
\cmidrule(lr){3-10}
 & & \multicolumn{2}{c}{\textbf{2B}} & \multicolumn{2}{c}{\textbf{4B}} & \multicolumn{2}{c}{\textbf{8B}} & \multicolumn{2}{c}{\textbf{32B}} \\
\cmidrule(lr){3-4}\cmidrule(lr){5-6}\cmidrule(lr){7-8}\cmidrule(lr){9-10}
 & & \textbf{Acc. (\%)} & \textbf{KV\% / GiB} & \textbf{Acc. (\%)} & \textbf{KV\% / GiB} & \textbf{Acc. (\%)} & \textbf{KV\% / GiB} & \textbf{Acc. (\%)} & \textbf{KV\% / GiB} \\
\midrule

\textbf{MMVU} & Baseline
& 39.94\nodrop & \kvsaving{0.00}{0.000}
& 52.03\nodrop & \kvsaving{0.00}{0.000}
& 56.64\nodrop & \kvsaving{0.00}{0.000}
& 64.45\nodrop & \kvsaving{0.00}{0.000} \\
\cline{2-10}

& Offset 5
& 39.94\drop{10.1} & \kvsaving{16.74}{0.233}
& \colorbox{red!1.8}{51.71}\drop{7.8} & \kvsaving{13.02}{0.233}
& 57.17\drop{8.3} & \kvsaving{13.02}{0.233}
& \colorbox{red!8.3}{62.96}\drop{4.4} & \kvsaving{7.33}{0.233} \\

& Offset 10
& \colorbox{red!2.4}{39.51}\drop{20.8} & \kvsaving{33.49}{0.466}
& \colorbox{red!6.5}{50.86}\drop{16.9} & \kvsaving{26.05}{0.466}
& \colorbox{red!9.5}{54.93}\drop{15.8} & \kvsaving{26.05}{0.466}
& \colorbox{red!4.7}{63.60}\drop{8.4} & \kvsaving{14.65}{0.466} \\

& Offset 15
& \colorbox{red!5.4}{38.97}\drop{33.6} & \kvsaving{50.23}{0.699}
& \colorbox{red!25.6}{47.43}\drop{28.5} & \kvsaving{39.07}{0.699}
& \colorbox{red!20.8}{52.89}\drop{25.0} & \kvsaving{39.07}{0.699}
& \colorbox{red!20.8}{60.71}\drop{15.7} & \kvsaving{21.98}{0.699} \\

\midrule

\textbf{MVBench} & Baseline
& 39.50\nodrop & \kvsaving{0.00}{0.000}
& 52.50\nodrop & \kvsaving{0.00}{0.000}
& 50.50\nodrop & \kvsaving{0.00}{0.000}
& 68.00\nodrop & \kvsaving{0.00}{0.000} \\
\cline{2-10}

& Offset 5
& \colorbox{red!22.2}{35.50}\drop{9.8} & \kvsaving{17.57}{0.244}
& \colorbox{red!19.4}{49.00}\drop{7.6} & \kvsaving{13.67}{0.244}
& 54.50\drop{8.3} & \kvsaving{13.67}{0.244}
& \colorbox{red!8.3}{66.50}\drop{4.6} & \kvsaving{7.69}{0.244} \\

& Offset 10
& \colorbox{red!50.0}{30.50}\drop{16.4} & \kvsaving{35.14}{0.488}
& \colorbox{red!16.7}{49.50}\drop{14.3} & \kvsaving{27.33}{0.488}
& 53.00\drop{14.6} & \kvsaving{27.33}{0.488}
& \colorbox{red!25.0}{63.50}\drop{8.3} & \kvsaving{15.38}{0.488} \\

& Offset 15
& \colorbox{red!5.6}{38.50}\drop{23.6} & \kvsaving{52.72}{0.731}
& \colorbox{red!2.8}{52.00}\drop{19.4} & \kvsaving{41.00}{0.731}
& 52.00\drop{17.9} & \kvsaving{41.00}{0.731}
& \colorbox{red!47.2}{59.50}\drop{10.8} & \kvsaving{23.06}{0.731} \\

\bottomrule
\end{tabular}
\end{adjustbox}
\end{table*}

Our findings in Section~\ref{sec:attention_dynamics} suggest that access to the original visual tokens becomes progressively less important along two dimensions: as autoregressive generation proceeds and in the later layers of the transformer. As attention weights may not be a direct measure of causal contribution to the final prediction, we test these observations through \emph{causal interventions} in which we propose to remove all visual tokens from the model's context at different points in generation time and transformer depth. These interventions directly verify our findings and test empirically whether continued access to the visual tokens is necessary for reasoning. If visual dependence indeed starts to diminish after the early window, then removing the original visual tokens afterwards should have only a limited effect on task performance. Similarly, as weak visual influence is observed in the final transformer layers, removing all visual tokens in those layers should lead to a limited effect on task performance. Beyond validating our findings, these interventions expose the practical trade-off between task accuracy and visual token reduction. Our causal intervention experiment is illustrated in Figure \ref{fig:method} for visual tokens removal across generation time (left) and across transformer depth (right). During attention in transformer models, only the keys and values are needed for past tokens, while the query is computed from the current token. Transformer models store the processed keys and values in a key-value (KV) cache.  Therefore, dropping visual tokens involves dropping their key-value cache, thereby also decreasing key-value cache memory during autoregressive decoding. 

\textbf{Dropping across generation time:} At a generation step $t_{\mathrm{drop}}$, we remove the KV cache entries of all $N$ visual tokens in $f_v^l$ from \textit{every} transformer layer $l \in \{1,\ldots,L\}$. For all subsequent reasoning tokens $g_y$ where $y \geq t_{\mathrm{drop}}$, the model can no longer attend to $f_v^l$ in any of the layers, and must continue generation using the layer-wise KV cache representations of previous text representations. 
\newline
\newline
\textbf{Dropping across transformer depth:} At layer $l_{\mathrm{drop}}$, we remove all $N$ visual tokens in
$f_v^{l_{\mathrm{drop}}}$ \textit{before} they enter that layer. As a result, no visual tokens are processed by layer $l_{\mathrm{drop}}$ or any subsequent layer ($l \geq l_{\mathrm{drop}}$). Since the tokens are discarded before entering these layers, no corresponding key-value (KV) cache entries are generated. We perform this during the prefill stage (before generation starts), and the dropping is maintained across the entire generation trajectory $y \in \{1,\ldots,Y\}$. Thus, layers $l < l_{\mathrm{drop}}$ operate normally and can still process visual information, whereas the upper layers $l \geq l_{\mathrm{drop}}$ operate without access to any visual tokens.
\newline
\newline
Experiments were performed using greedy decoding without sampling, with a maximum generation length of 2,048 new tokens. Each experiment was performed on a single NVIDIA H200 GPU with exactly the same hardware, and we used multiple GPUs in parallel to speedup experiments. For positional encoding, retained tokens preserved their previously computed positional information, without recomputing or reassigning positional embeddings in the sequence.

\section{Results on Image Benchmarks}
\subsection{Dropping Across Time}

We evaluate four dropping times: $t_{\mathrm{drop}} \in \{50, 100, 200, 300\}$. Dropping performance across those generation times is reported in Table \ref{tab:dropping-generation-time} for all benchmarks and models. For each benchmark-model combination, we report the baseline performance without any dropping. We also report in \textcolor{green!50!black}{green} the prompt token reduction (\%). The prompt token reduction \% represents the percentage of visual tokens removed from the \textit{full input prompt} (including both visual and textual prompt tokens), averaged across all samples. We measure this as $\frac{N}{N+T}\times 100\%$, where $N$ is the number of visual tokens and $T$ is the number of textual prompt tokens. We measure the prompt reduction ratio because it represents how much tokens are removed from the full KV cache of the model, rather than a subset of it. We also impose two criteria when calculating the prompt token reduction (\%): (1) the number of generated tokens is greater than the drop step, and (2) the number of generated tokens is less than or equal than that of the baseline. We use those two requirements because visual token dropping is considered ineffective when the generation terminates before the drop step (e.g., generation terminates at step 150 when $t_{drop}=200$). Furthermore, if dropping visual tokens leads to a longer generation than the baseline, then the model may have generated additional reasoning tokens to compensate for the missing visual information, making it difficult to isolate the direct influence of the original visual tokens on task performance. Samples not satisfying \textit{either} of the following conditions are assigned a prompt token reduction of 0\%. These two conditions ensure a fair comparison with the baseline by assigning zero reduction whenever dropping is not actually applied or leads to a higher token count, as

We also report and analyze the number of generated tokens in Section \ref{app:num_generated_tokens}. We also summarize the results of Table \ref{tab:dropping-generation-time} as a graph in Figure \ref{fig:img_kv_cache_drop}, where we show how performance varies as $t_{drop}$ increases. For each point, we also provide the prompt token reduction \%.

The results across all LVLMs consistently show that dropping visual tokens early degrades performance, causally verifying a strong dependence on visual tokens in the early stage of generation. As $t_{drop}$ increases (i.e., as visual tokens are removed later during generation), performance gradually recovers. Around timestep 200, the performance degradation becomes minimal, suggesting that visual dependence is significantly reduced. We therefore identify this point as the \textit{general best trade-off} point, providing a favorable balance between preserving performance and reducing visual tokens across most benchmarks. At this point, the average prompt token reduction ratio is around 22\%. For MMStar and HallusionBench, however, the early visual dependence window appears to be shorter, with performance fully recovering as early as timestep 100. At this point, the average prompt token reduction ratio is around 30\%. Overall, these causal intervention results are consistent with the attention dynamics observed in Section~\ref{subsection:dynamics_generation_time}, which confirms our findings. The stronger performance degradation observed on WeMath and LogicVista under early visual token removal suggests a stronger dependence on visual information during the initial stages of reasoning. This may be attributed to the multi-step nature of mathematical and logical reasoning, where intermediate reasoning steps often require continued visual access, making them more sensitive to early visual token removal. In contrast, HallusionBench and MMStar show greater robustness to visual token removal, potentially because their tasks rely more heavily on an initial understanding of the image, with subsequent reasoning requiring less sustained access to visual information. 

\subsection{Dropping across Depth} We evaluate three different layer dropping indices: $l_{\mathrm{drop}} \in \{L-5, L-10, L-15\}$, where $L$ is the total number of layers. For example, $L-5$ indicates that we apply dropping 5 layers before the final layer, where 5 is the offset. Dropping performance across those layer indices is reported in Table \ref{tab:image_token_drop} for all benchmarks and models, along with the baseline performance without any dropping. We also report in \textcolor{green!50!black}{green} the all-layer visual token reduction (\%). The all-layer visual token reduction (\%) represents the percentage of visual tokens dropped \textit{across all layers} of the transformer model, averaged across all samples, with ineligible samples contributing 0\% reduction. The all-layer visual token reduction ration for an eligible sample is measured as $\frac{N(L-l_{\mathrm{drop}})}{NL}\times 100\%=\frac{L-l_{\mathrm{drop}}}{L}\times 100\%$. For the same reason discussed in the previous experiment, samples for which the number of generated tokens exceeds that of the baseline are assigned a all-layer visual token reduction of 0\%. The number of generated tokens are reported and analyzed in Section \ref{app:num_generated_tokens}. We also summarize the results of Table \ref{tab:image_token_drop} as a graph in Figure \ref{fig:img_token_drop}, where we show how performance varies as the offset increases. For each point, we also provide the all-layer visual token reduction \%.

Results in Table~\ref{tab:image_token_drop} and Figure~\ref{fig:img_token_drop} show that all visual tokens can be removed from the upper transformer layers without substantially affecting reasoning performance. This finding is consistent across all LVLMs. Dropping visual tokens from the last five layers (Offset 5) produces a very slight degradation across most benchmarks and model sizes (and in many times, improves), while achieving up to 12\% all-layer visual token reduction across all layers, indicating that the final layers contribute relatively little direct visual processing and mainly operate on representations processed by lower layers. However, extending the removal to earlier layers causes a consistent performance decline, suggesting that intermediate layers are important for multimodal reasoning. Overall, the causal intervention results agree with the layer-wise attention analysis in Section \ref{subsec:attention_dynamics_across_transformer_depth}, and we can remove visual tokens near the top of the transformer stack. 

\begin{figure*}
    \centering
    \includegraphics[width=1\linewidth]{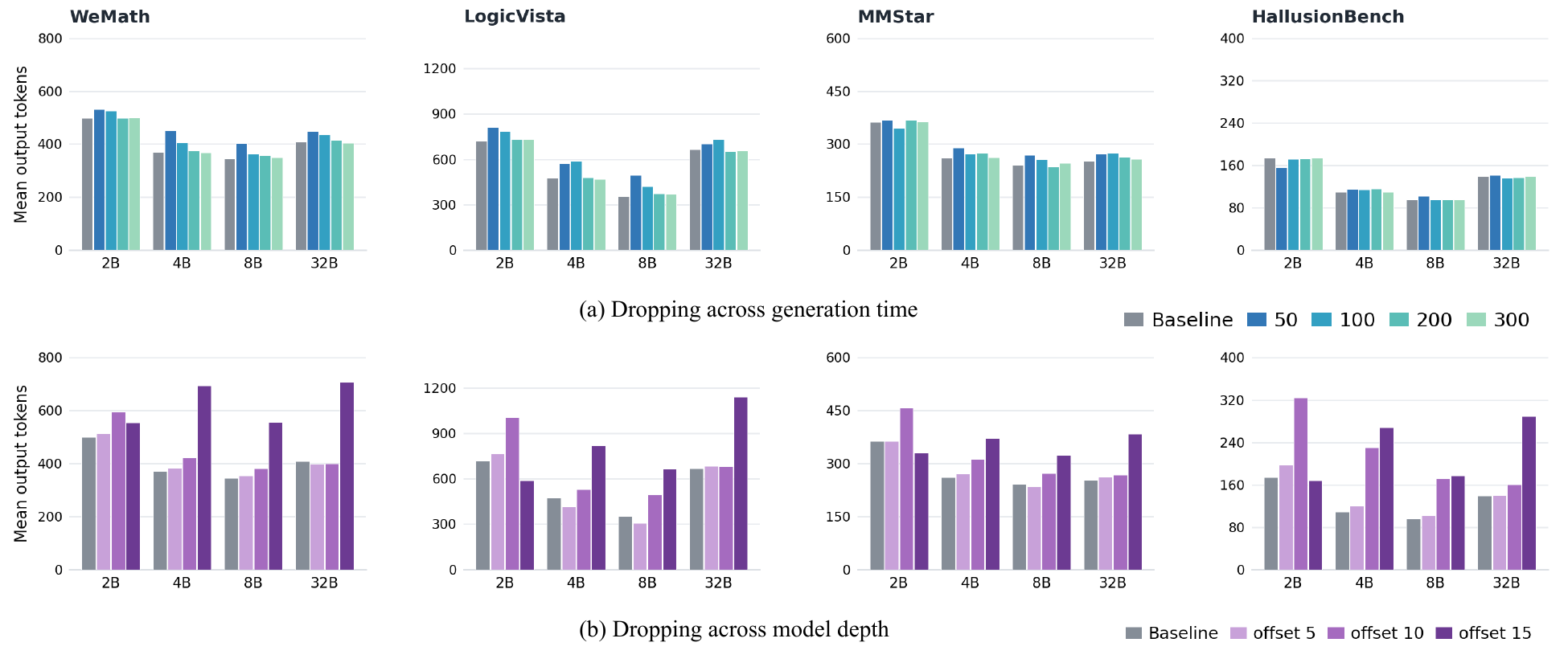}
    \caption{The average token generation length across all samples for (a) dropping visual KV cache across generation time, and (b) dropping image tokens across model depth, for WeMath, LogicVista, MMStar and HallusionBench}
    \label{fig:num_tokens_generated_imgs}
\end{figure*}

\begin{figure*}
    \centering
    \includegraphics[width=1\linewidth]{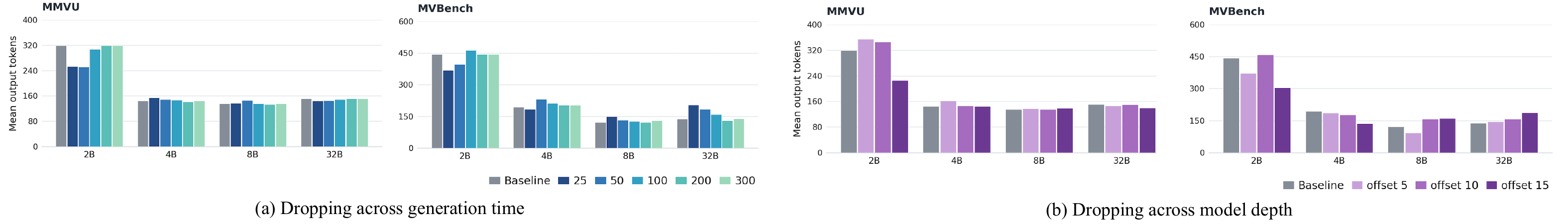}
    \caption{The average token generation length across all samples for (a) dropping visual KV cache across generation time, and (b) dropping image tokens across model depth, for MMVU and MVBench.}
    \label{fig:num_tokens_generated_video}
\end{figure*}

\section{Results on Video Benchmarks}
\label{app:sec_video_causal_intervention_results}

\subsection{Dropping Across Time:} We report performance results for dropping across time for video benchmarks MMVU and MVBench, along with the prompt token reduction percentages. Since video benchmarks generally involve shorter generated responses than image benchmarks (see Figure \ref{fig:num_tokens_generated_video}), many sequences terminate before reaching later dropping steps, limiting their applicability. We therefore additionally evaluate an earlier dropping step at timestep 25. We also evaluate the decoder KV-cache memory savings achieved by KV Cache dropping. Only samples whose generation length exceeds the dropping step are considered. For those samples, we compute the memory savings at the drop point, where both caches contain $t_{drop}$ generated tokens. For a single token at a single layer, the number of bytes for both keys and values are $2 H_{kv} d_h b$, where $H_{kv}$ are the number of key-value heads and $d_h$ is the dimensionality of each head $h$, and $b=2$ bytes per element for BF16 mixed precision format. We report the absolute savings in GiB ($1\,\mathrm{GiB}=1024^3$ bytes). We first report the KV-cache reduction percentage. The KV-cache reduction percentage is calculated as the total memory saved divided by the total pre-dropping KV-cache memory across eligible samples, multiplied by 100. For example, if the KV cache memory before dropping is 531 MB, and after dropping becomes 35 MB, then the reduction percentage is around 93\%. We also report the maximum absolute memory saved across eligible samples, reporting an upper bound on the savings achieved. 

Results for token dropping across generation time are presented in Table \ref{tab:dropping-generation-time-video-benchmarks}. We report values separately for each model, benchmark, and dropping step. Our results suggest that the early visual dependance window is shorter for video inputs than for images. Remarkably, dropping at step 50 can eliminate over 65\% of prompt tokens and approximately 98\% of the KV cache at the drop step, without degrading accuracy. This indicates that the visual dependence window is earlier in videos, allowing substantial KV cache savings.

\subsection{Dropping Across Depth:} We report performance results for dropping across depth for video benchmarks MMVU and MVBench, along with the all-layer visual token reduction percentages. Note that for dropping across depth, there are no KV caches created in the first place, since the image tokens are dropped \textit{before} entering a layer. Therefore, to calculate the memory savings, we must consider the counterfactual scenario in which image tokens are not dropped, in which case KV caches would be created for the last $(L-\text{offset})$ layers. We report the prompt-only KV-cache memory savings. That is the KV cache memory saved during prefil, and is calculated as the KV cache memory saved by dropping image tokens, divided by the total KV cache memory that would be created if tokens were not dropped. We report this as it isolates the prefill KV-cache footprint, independent of generated answer length, and makes comparisons across benchmarks, model sizes, and dropping offsets more consistent. Note also that this memory savings persist throughout the full generation trajectory since the dropping occurs at the prefill stage. We also report the maximum memory saved across all samples.

Results for token dropping across depth are presented in Table \ref{tab:image_token_drop_video}. Removing visual tokens from the final five transformer layers generally preserves performance, particularly on MMVU. However, extending removal to earlier layers often causes substantial degradation, suggesting that intermediate layers remain important for temporal and visual reasoning, mirroring the results for image benchmarks. Note also that, because dropping begins at layer \(L-\text{offset}\), the earlier layers still store KV tensors for all prompt tokens. For this reason, the savings are usually lower than dropping across time. 
\newline
\newline
\textbf{Latency:} Regarding latency, we observe no speedups or improvements in throughput despite dropping image tokens. We attribute this to two main factors. The first factor concerns GPU utilization and parallelism. Modern high-performance GPUs can already process large token sequences efficiently through highly optimized parallel operations. Dropping tokens may decrease the available parallelism and GPU utilization, limiting speedups since fewer tokens achieve less parallel work. This makes GPU execution less efficient, and may even increase latency. The second factor concerns the distribution of computational costs during inference. In multimodal LLMs, inference latency is often dominated by (1) the prefill stage, where the number of image or video tokens substantially exceeds the number of generated text tokens, and (2) the computationally expensive MLP layers, which account for approximately 70\% of the LLM's computational workload. Although dropping image tokens reduces the computational cost of attention during autoregressive decoding, it does not reduce the cost of the MLP operations or the initial prefill stage. As a result, the overall latency improvements remain limited. Importantly, our causal interventions are not intended as an efficient inference method or a technique for accelerating inference. Instead, their primary purpose is to causally verify how visual dependence evolves across generation time and transformer depth during multimodal reasoning, characterize its temporal and depth-wise distribution, and where direct access to visual information remains necessary.

\section{Number of Generated Tokens}
\label{app:num_generated_tokens}
In Figure \ref{fig:num_tokens_generated_imgs}, we analyze how the generation length changes in image benchmarks as the visual KV cache is dropped across the generation time (a) and as image tokens are dropped across the transformer depth (b). We report the average number of generated tokens across all samples for all four benchmarks: WeMath, LogicVista, MMStar, and HallusionBench. Each group of bars shows Qwen3-VL-2B, 4B, 8B, and 32B. Grey bars indicate the no-drop baseline; colored bars indicate the drop settings. Depth drop at offset $k$ means dropping at layer $L-k$, where $L$ is the final decoder layer index. Each benchmark uses the same vertical scale across its time and depth panels. For both dropping across time and depth, most of the generation length stays roughly the same as the baseline. The biggest changes generally occur when dropping visual tokens across very early depths (15 layers earlier). For example, on HallusionBench, Qwen3-VL-4B increases from an average of 109 baseline tokens to 268 at offset 15, approximately +146\%. This increase is accompanied by a decrease in the HallusionBench score from 74.55 to 68.03 (Table~\ref{tab:image_token_drop}), indicating a deterioration in hallucination robustness. These findings suggest that removing visual information at earlier layers can lead to longer yet less visually grounded responses. Without access to the removed visual tokens, the model may increasingly rely on visual information encoded in earlier hidden representations or on its learned language priors, potentially producing additional reasoning and hallucinated content to compensate for the reduced visual grounding. 

We also report in Figure \ref{fig:num_tokens_generated_video} the average number of tokens generated for video benchmarks MMVU and MVBench, across generation time (a) and model depth (b). For video benchmarks, most of the generated token count stays roughly the same as the baseline.

\section{Conclusion}

We investigated how visual information is utilized during long-form LVLM reasoning. Through attention analysis and causal interventions, we show that visual influence is concentrated within an early generation window and intermediate transformer layers, while later reasoning increasingly relies on textual representations. These findings reveal that continuous access to all visual tokens is not always necessary and suggests a time and depth-wise pattern in the use of visual information and a pattern of visual reliance in LVLMs.

\clearpage
\section{Prompts}
\label{app:prompts}
We provide the prompt used for all our benchmarks in Table \ref{tab:evaluation_prompts}. 

\begin{table*}
\centering
\caption{Prompt templates used for evaluation on the four multimodal benchmarks.
Dataset-specific content is represented using curly-braced placeholders.
The prompts were kept fixed for all samples within each benchmark.}
\label{tab:evaluation_prompts}
\small
\setlength{\tabcolsep}{5pt}
\renewcommand{\arraystretch}{1.15}
\begin{tabularx}{\textwidth}{p{2.3cm} p{3.0cm} X}
\toprule
\textbf{Benchmark} &
\textbf{Task / Output} &
\textbf{Prompt} \\
\midrule

\textbf{HallusionBench}
&
Visual yes/no question.
Output: \texttt{yes} or \texttt{no}.
&
Now, we require you to solve a yes/no question shown in the image.
Please briefly describe your thought process and provide the final answer
(yes or no).

\textbf{Question:} \texttt{\{question\}}

Regarding the format, please answer following the template below, and be
sure to include two \texttt{<>} symbols:

\texttt{<Thought process>: «your thought process» <Answer>: «your option»}
\\
\midrule

\textbf{WeMath}
&
Visual Multiple-choice mathematical reasoning.
Output: option.
&
Now, we require you to solve a multiple-choice math question.
Please briefly describe your thought process and provide the final answer
(option).

\textbf{Question:} \texttt{\{question\}}

\textbf{Option:} \texttt{\{options\}}

Regarding the format, please answer following the template below, and be
sure to include two \texttt{<>} symbols:

\texttt{<Thought process>: «your thought process» <Answer>: «your option»}
\\
\midrule

\textbf{LogicVista}
&
Visual multiple-choice reasoning.
Output: option.
&
Now, we require you to solve a multiple-choice question shown in the image.
Please briefly describe your thought process and provide the final answer
(option).

\textbf{Question:} \texttt{\{question\}}

Regarding the format, please answer following the template below, and be
sure to include two \texttt{<>} symbols:

\texttt{<Thought process>: «your thought process» <Answer>: «your option»}
\\
\midrule

\textbf{MMStar}
&
Visual Multiple-choice multimodal reasoning.
Output: option.
&
Now, we require you to solve a multiple-choice question.
Please briefly describe your thought process and provide the final answer
(option).

\textbf{Question:} \texttt{\{question\}}

Regarding the format, please answer following the template below, and be
sure to include two \texttt{<>} symbols:

\texttt{<Thought process>: «your thought process» <Answer>: «your option»}
\\
\midrule

\textbf{MMVU}
&
Video question answering.
Output: answer or option letter.
&
Now, we require you to answer the question for the video.
Please briefly describe your thought process and provide the final answer.
In the case of multiple-choice questions, only answer with the correct option letter.

\textbf{Question:} \texttt{\{question\}}

Regarding the format, please answer following the template below, and be
sure to include two \texttt{<>} symbols:

\texttt{<Thought process>: «your thought process» <Answer>: «your answer»}
\\
\midrule

\textbf{MVBench}
&
Video multiple-choice reasoning.
Output: option.
&
Now, we require you to solve a multiple-choice math question.
Please briefly describe your thought process and provide the final answer
(option).

\textbf{Question:} \texttt{\{question\}}

Regarding the format, please answer following the template below, and be
sure to include two \texttt{<>} symbols:

\texttt{<Thought process>: «your thought process» <Answer>: «your option»}
\\

\bottomrule
\end{tabularx}
\end{table*}

\clearpage
\bibliographystyle{IEEEtran}
\bibliography{references}

\end{document}